\documentclass[times,twocolumn,final,authoryear]{elsarticle}

\usepackage{prletters}
\usepackage{framed,multirow}

\usepackage{amssymb}
\usepackage{latexsym}
\usepackage{amsmath}
\usepackage{url}
\usepackage{xcolor}
\definecolor{newcolor}{rgb}{.8,.349,.1}

\usepackage[percent]{overpic}
\usepackage{algorithm}
\usepackage{algorithmic}
\usepackage{makecell}
\DeclareMathOperator*{\argmax}{argmax}

\journal{Pattern Recognition Letters}

\begin{document}

\thispagestyle{empty}

\clearpage

\ifpreprint
  \setcounter{page}{1}
\else
  \setcounter{page}{1}
\fi

\begin{frontmatter}

\title{Video Semantic Object Segmentation by Self-Adaptation of DCNN}

\author{Seong-Jin \snm{Park}\corref{cor1}} 
\cortext[cor1]{Corresponding author: 
  Tel.: +82-54-279-2881;  }
\ead{windray@postech.ac.kr}
\author{Ki-Sang \snm{Hong}}

\address{Department of Electrical Engineering, POSTECH, Namgu Pohang, Republic Korea}

\received{1 May 2013}
\finalform{10 May 2013}
\accepted{13 May 2013}
\availableonline{15 May 2013}
\communicated{S. Sarkar}

\begin{abstract}
This paper proposes a new framework for semantic segmentation of objects in videos. We address the label inconsistency problem of deep convolutional neural networks (DCNNs) by exploiting the fact that videos have multiple frames; in a few frames the object is confidently-estimated (CE) and we use the information in them to improve labels of the other frames. Given the semantic segmentation results of each frame obtained from DCNN, we sample several CE frames to adapt the DCNN model to the input video by focusing on specific instances in the video rather than general objects in various circumstances. We propose offline and online approaches under different supervision levels. In experiments our method achieved great improvement over the original model and previous state-of-the-art methods.
\end{abstract}


\end{frontmatter}


\section{Introduction}

Semantic segmentation assigns all pixels in an image to semantic classes; it gives finely-detailed pixel-level information to visual data and can build a valuable module for higher-level applications such as image answering, event detection, and autonomous driving. Conventional semantic segmentation techniques for images have been mostly built using handcrafted features on conditional random fields (CRFs) \citep{russell2009associative,ladicky2010and, yao2012describing}. Recently deep convolutional neural networks (DCNNs) have achieved great success in classification \citep{krizhevsky2012imagenet, simonyan2014very, szegedy2015going,he2015deep}, so they have been widely applied to semantic segmentation approaches \citep{long2015fully, chen2014semantic,noh2015learning,zheng2015conditional,lin2015efficient,liu2015semantic,yu2015multi}. 
\begin{figure}[ht]\centering
\begin{tabular}{cccccc}
\includegraphics[width=0.88in]{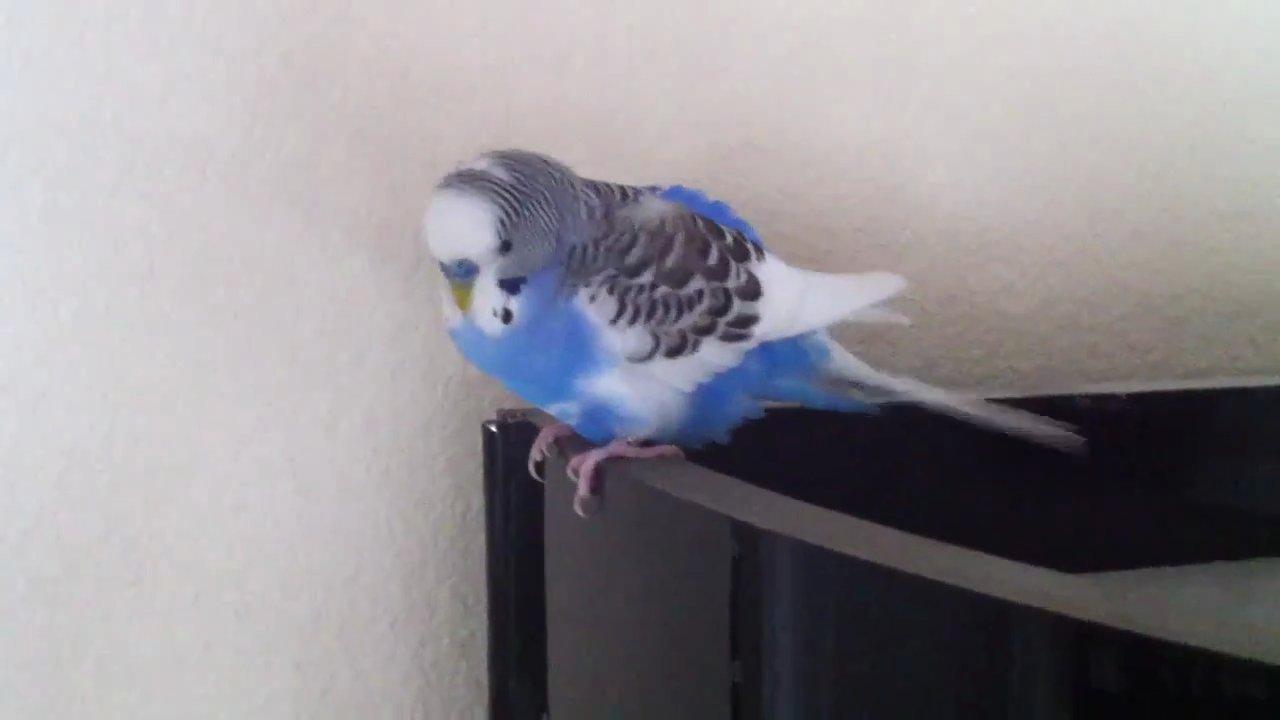}
\includegraphics[width=0.88in]{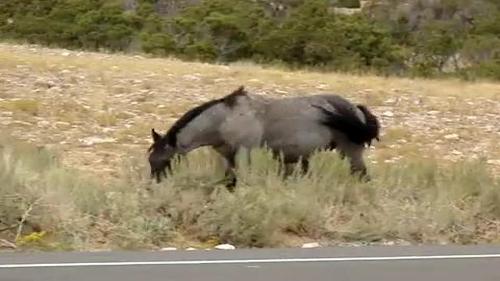}
\includegraphics[width=0.88in]{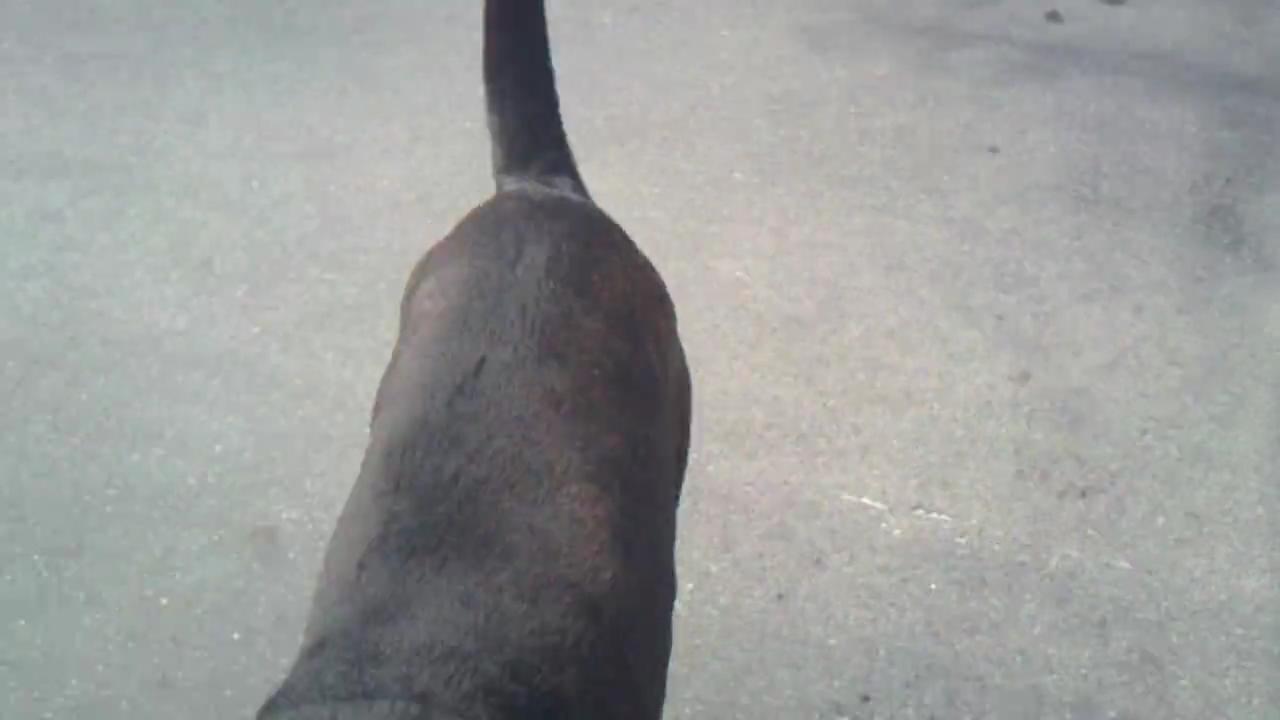}\\
\includegraphics[width=0.88in]{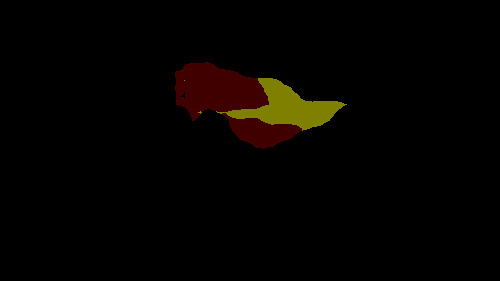} 
\includegraphics[width=0.88in]{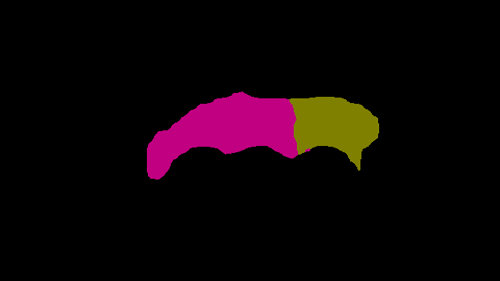} 
\includegraphics[width=0.88in]{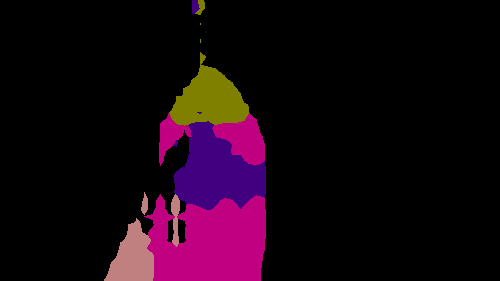} \\
(a)\\
\includegraphics[width=0.88in]{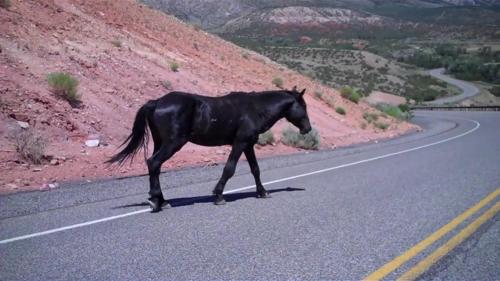}
\includegraphics[width=0.88in]{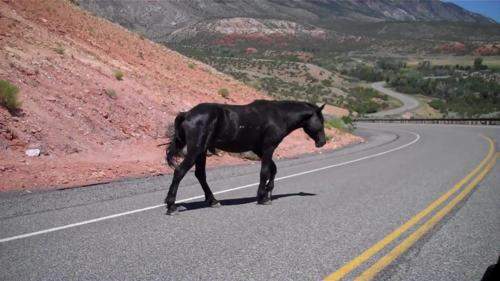}
\includegraphics[width=0.88in]{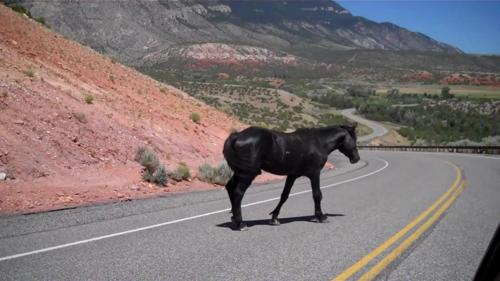}\\
\includegraphics[width=0.88in]{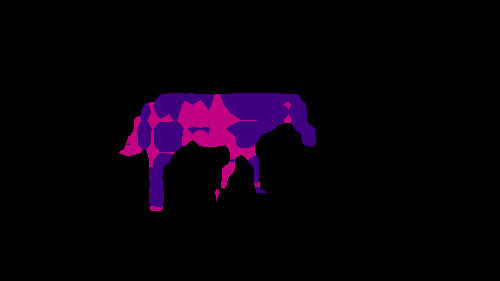}
\includegraphics[width=0.88in]{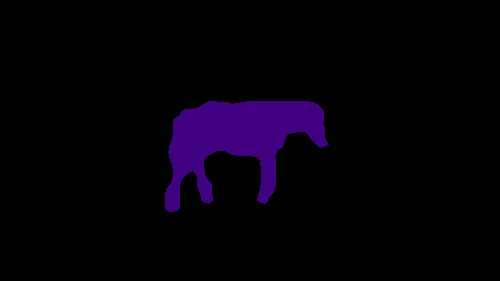} 
\includegraphics[width=0.88in]{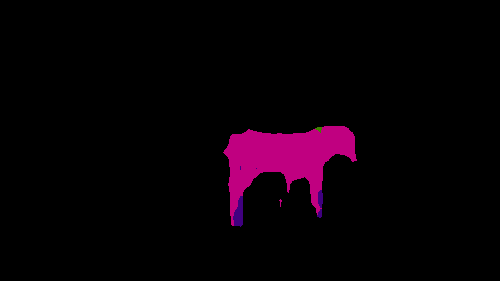} \\
(b)
\end{tabular} 
\caption{Problems when the pre-trained DCNN model is applied directly to a video frame (top: a frame, bottom: the result). Different colors of results represent different classes. (a) Objects segmented into different classes. (b) Label wavers between visually-similar categories (from left to right: frame 1 - mixed, frame 31 - dog, frame 61 - horse). {\bf Best viewed in color.}}
\end{figure}

A video consists of a sequence of images, so image-based models can be applied to each of them. For instance, there was an attempt to apply object and region detectors pre-trained on still images to a video \citep{zhang2015semantic}; they used pre-trained object detectors to generate rough object region proposals in each frame. Similarly, we adopt an approach that employs an image-trained model to process a video, but instead of a conventional object detector, we apply a DCNN semantic segmentation model to each frame. However, the DCNN model can show spatially inconsistent labels as previously reported in \citep{qi2015semantic} when it is applied to an image. This inconsistency is exacerbated for video due to various factors such as motion blur, video compression artifacts, and sudden defocus \citep{kalogeiton2015analysing}. When the model is applied directly to video, the labeling result for an object can be segmented into different classes, and can waver between visually confusing categories among frames (Fig. 1). Human vision also experiences such difficulty of recognition under certain circumstances. Our framework is motivated by the following question: `How does a human recognize a confusing object?' When a human has difficulty identifying the object, she can guess its identity by using learned models and focusing on the object for a while. Then, she can recognize the object by referring to the moments during which it is unambiguous; i.e., she tunes her model to the specific object while regularizing small appearance changes. In a way analogous to this process, our framework takes advantage of multiple frames of a video by emphasizing confidently-estimated (CE) frames to adapt a pre-trained model to a video.

\begin{figure}[t]
\begin{center}
\begin{overpic}[width=1.00\linewidth]{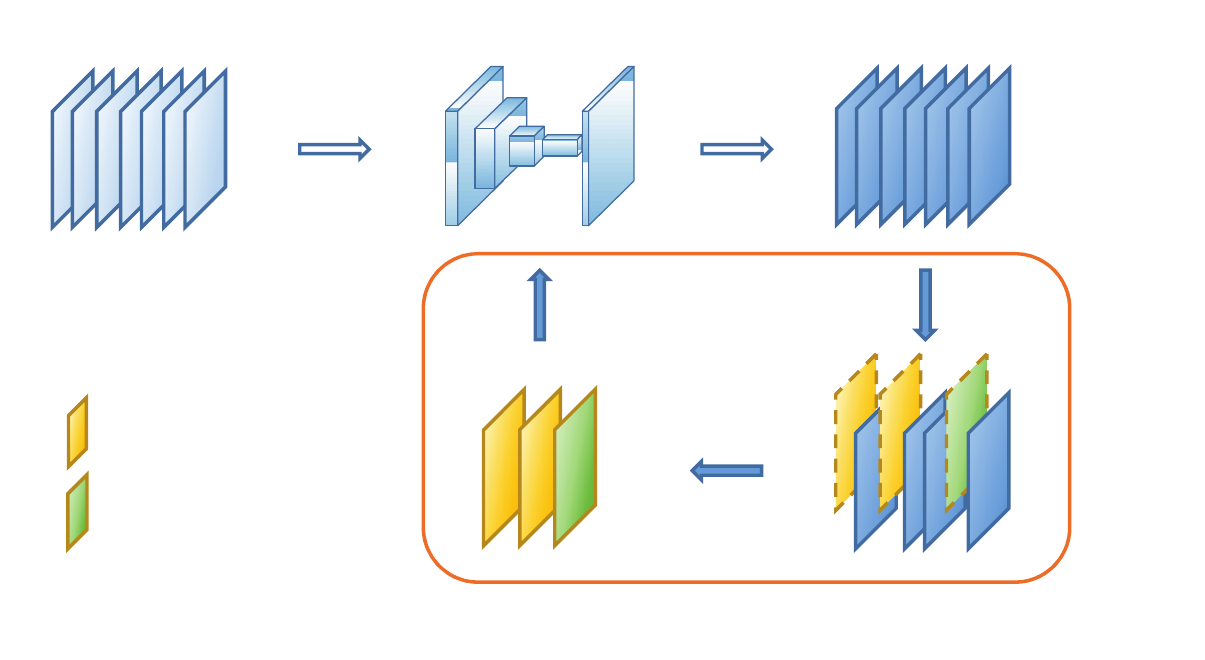} 
\put(2,52){\footnotesize{Video frames}}
\put(8.8,18){\footnotesize{Globally confident}}
\put(8.8,11.5){\footnotesize{Locally confident}}
\put(24,52){\footnotesize{DCNN Semantic Segmentation}}
\put(72,52.5){\footnotesize{Output}}
\put(79,28.5){\footnotesize{Select frames}}
\put(47,28.5){\footnotesize{Adapt model}}
\put(54,17){\footnotesize{Set labels}}
\put(40,3){\footnotesize{{\bf Generate self-adapting dataset}}}

\end{overpic}
\end{center}
\caption{Main framework of our method. {\bf Best viewed in color.}}
\end{figure}
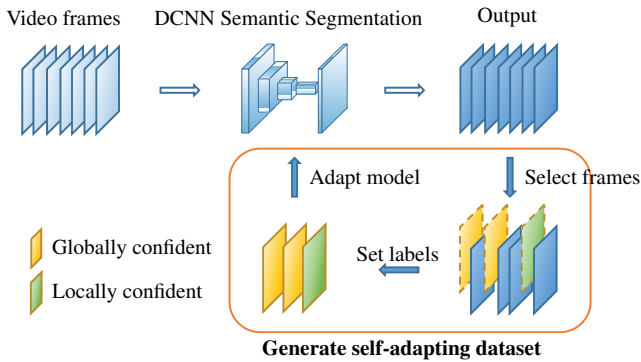
The key idea of our method is to propagate the belief of CE frames to the other frames by fine-tuning DCNN model; we apply a pre-trained DCNN model to each frame to guess the object's class, and collect frames in which the estimation is globally confident or locally confident. Then, we use the set of CE frames as a training set to fine-tune the pre-trained model to the instances in a video (Fig. 2). We restrict the DCNN model to be video-specific rather than general purpose; i.e., we make the model focus on the specific instances in each video. In our procedures, we only use the label of CE regions, and let the uncertainly-estimated (UE) regions be determined by the CE frames. We also incorporate weak labels (i.e., manually-annotated class labels of objects in the video) to prevent a few incorrect labels from degrading the model. Our procedures to generate a self-adapting dataset and to use CE frames to update the model can recover the uncertain or misclassified parts of UE frames that include multiple objects.

We also propose an online approach that is implemented in a way similar to object tracking, because object tracking and online video object segmentation are closely related in that both tasks should trace the appearance change of an object while localizing it. Then we combine the batch and online results to improve the motion-consistency of segmentation.

We validate our proposed method on the Youtube-Object-Dataset \citep{prest2012learning,jain2014supervoxel, zhang2015semantic}. In experiments our model greatly improved the pre-trained model by mitigating its drawback even when we do not use the weak labels.
\section{Related work}
Recent image semantic segmentation techniques have been propelled by the great advance of DCNN for the image classification task \citep{krizhevsky2012imagenet, simonyan2014very, szegedy2015going,he2015deep}. Based on the classification network, \cite{long2015fully} extended a convolutional network to a fully-convolutional end-to-end training framework for pixel-wise dense prediction. \cite{chen2014semantic} used a hole algorithm to efficiently compute dense feature maps, and combined the output of the network into a fully-connected CRF. Several follow-up studies proposed more-sophisticated combinations of CRF framework with DCNNs \citep{zheng2015conditional,lin2015efficient,liu2015semantic}. \cite{yu2015multi} proposed a modified architecture that aggregates multi-scale context by using dilated convolutions specifically designed for dense prediction.

Due to the difficulty of pixel-wise annotation for frames, most video semantic object segmentation techniques have been built on a weakly-supervised setting that is given only video-level class labels of objects appearing in a video. 
\cite{hartmann2012weakly} trained weakly-supervised classifiers for several spatiotemporal segments and used graphcuts to refine them. \cite{tang2013discriminative} determined positive concept segments by using a concept-ranking algorithm to compare all segments in positive videos to the negative videos. \cite{liu2014weakly} proposed a label transfer scheme based on nearest neighbors. The natural limitation of the weakly-supervised approach is that it has no information about the location of target objects. Because a video is a sequence of images, \cite{zhang2015semantic} used object detectors that had been pre-trained on still images, and applied them to a video to localize object candidates; in each frame the method generates several object proposals by using object detectors that correspond to the given labels and by using rough segmentation proposals based on objectness. Although the object detection gives the spatial information as a bounding box around objects that have a semantic class label, the information is not sufficient for pixel-wise segmentation. Thus we use a DCNN semantic segmentation model pre-trained on images to give the pixel-wise spatial extent of the object and its semantic label at the same time, and adapt the image-based DCNN model to the input video. In contrast to most existing approaches that focus on modeling temporal consistency at pixel or region levels, our framework does not necessarily assume that the neighboring frames should be similar, and because it samples several frames that may capture different appearances of an object, our framework is relatively insensitive to sudden changes of the object.

Another related topic is video object segmentation with semi-supervised video \citep{ali2011flowboost,ramakanth2014seamseg,tsai2012motion,badrinarayanan2010label,jain2014supervoxel, fathi2011combining}, which is given pixel-wise annotation of certain frames. Especially, \cite{jain2014supervoxel} proposed a method that employs supervoxels to overcome the myopic view of consistency in pairwise potentials by incorporating additional supervoxel-level higher-order potential. \cite{fathi2011combining} developed an incremental self-training framework by iteratively labeling the least uncertain frame and updating similarity metrics. The framework is similar to ours in that we update a model based on previously-estimated frames, although we neither assume pixel-wise annotation nor require superpixels, which often wrongly capture the boundary of the object. We also update the pre-trained DCNN model, in contrast to the method of \cite{fathi2011combining} that updates a simple similarity metric based on hand-crafted features such as SIFT and a color histogram.

\section{Method: AdaptNet}

We assume that a video includes at least a few CE frames and that they are helpful to improve the results of the UE frames. The main idea of our method is to propagate the belief of those CE frames by fine-tuning DCNN model. Thus our main framework consists of the following steps: selection of CE frames, label map generation, and adaptation of the model to the input video. We describe the detailed algorithm of these steps in the following subsections. We first propose a batch (offline) framework, then an online framework. We combine those two results to improve the motion-consistency of segmentation by incorporating optical flow. At the end of the section, we mention simple extensions to the unsupervised batch algorithm.
\subsection{Batch}

Let $\mathcal{F}$ denote a set of frame indices, and $\mathcal{W}$ denote a set of given weak labels of the input video. We begin by applying a pre-trained DCNN model $\theta$ to each frame $f\in \mathcal{F}$, then use softmax to compute the probability $P(x_i|\theta)$ that the $i$-th pixel is a member of each class $x_i\in\mathcal O$, where $\mathcal O$ denotes the set of object classes and background. The semantic label map $S$ can be computed using the argmax operation for every $i$-th pixel: $S(i)=\arg\max_{x_i} P(x_i|\theta)$. 

To adapt the DCNN model $\theta$ to the input video, we collect a self-adapting dataset $\mathcal{G}$ that consists of CE frames and corresponding label maps. We collect globally-CE and locally-CE frames and compute the respective label maps $G^g$ and $G^l$ to construct the self-adapting dataset. The procedures to select the frames and to compute the labels are  described in the following and summarized in Algorithm 1. We first perform connected-component analysis on each class map of $S$ to generate a set $\mathcal{R}$ of object candidate regions. For each $k$-th segmented label map $R_k\in \mathcal{R}$ we measure the confidence $C(R_k)$ of the estimated regions, where the $C(\cdot)$ operator takes a label map as input and computes the average probability that the pixels labeled as objects in the label map have the corresponding class labels.

  Then we generate the label map $G^g_f$ by setting the label of the region only when its confidence exceeds a high threshold $t_o$. We also set the background label for every pixel for which the probability $P(x_i=bg|\theta)$ of being background exceeds threshold $t_b$. 
  
To complete $G^g_f$, the remaining uncertain regions must be processed. For this purpose, we let the remaining pixels have the {\it ``ignored''} label. The uncertain {\it ``ignored''} pixels are not considered during computation of the loss function for model update. We also ignore all pixels that have labels that are not in the set $\mathcal{W}$ of given weak labels of a video. We add globally-CE frames with $G^g_f$ that has at least one confident region (i.e., $C(G^g_f)>0$) to the self-adapting dataset $\mathcal{G}$. 
  
Because the selected frames might be temporally unevenly distributed, the model can be dominated by frames that are selected during a short interval. To mitigate the resulting drawback and regularize the model, we also select the locally-CE frames that have best object confidences during every period $\tau_b$ although the frames do not include globally-CE object regions. We determine the locally-CE frame and its label map $G^l$ as follows: we generate a label map $G^l_f$ for every frame $f$ by keeping the label of all pixels only if the label $S(i)$ is included in $\mathcal{W}$, while setting the background as before. We measure the confidence of a frame by computing $C(G^l_f)$ and we regard as locally-CE the frame that has the highest confidence during every section of $\tau_b$ frames. If the locally-CE frame is not already selected as a globally-CE frame, we add it to the self-adapting dataset $\mathcal{G}$.
\begin{algorithm}[t]
 \caption{AdaptNet-Batch} \label{Alg1}
\begin{algorithmic}[1]
\STATE Given: DCNN model $\theta$, a set of weak labels $\mathcal{W}$
\STATE Local best confidence $d=0$
\STATE {\bf for} $f \in \mathcal{F}$ {\bf do}
\STATE ~~~~Initialize $G^g_f, G^l_f$ to "ignored" labels
\STATE ~~~~Compute $P({\bf x}|\theta)$ and $S=\arg\max_{\bf x}P({\bf x}|\theta)$
\STATE ~~~~Compute set $\mathcal{R}$ of connected components in $S$
\STATE ~~~~{\bf for} $R_k \in \mathcal{R}$ {\bf do}
\STATE ~~~~~~~~{\bf if} $S(i) \notin \mathcal{W}, i \in R_k$ {\bf then} {\bf continue}
\STATE ~~~~~~~~{\bf if} $C(R_k)>t_o$ {\bf then} Set $G^g_f(i)=S(i), \forall i \in R_k$
\STATE ~~~~~~~~Set $G^l_f(i)=S(i), \forall i \in R_k$
\STATE ~~~~Set $G^g_f(i)=G^l_f(i)=0, \forall i, s.t. P(x_i=bg|\theta)>t_b$
\STATE ~~~~{\bf if} $C(G^g_f)>0$ {\bf then} $\mathcal{G}\leftarrow \mathcal{G}\cup\{G^g_f\}$
\STATE ~~~~{\bf if} $C(G^l_f)>d$ {\bf then}
\STATE ~~~~~~~~Update $t = f$ and $d=C(G^l_f)$
\STATE ~~~~{\bf if} $f$ mod $\tau_b = 0$ {\bf then}
\STATE ~~~~~~~~{\bf if} $G^g_t \notin \mathcal{G}$ {\bf then}
\STATE ~~~~~~~~~~~~$ \mathcal{G}\leftarrow \mathcal{G}\cup\{G^l_t\}$
\STATE ~~~~~~~~Initialize $d=0$
\STATE Finetune DCNN model $\theta$ to $\theta'$ using the set $\mathcal{G}$
\end{algorithmic}
\end{algorithm}

Given the self-adapting dataset $\mathcal{G}$ constructed from the above procedures, we finally adapt the DCNN model $\theta$ to the video by fine-tuning the model to $\theta'$ based on the dataset. Then, we compute the new label map by applying $\theta'$ to every frame.

\begin{figure*}[ht]
\begin{center}
\begin{overpic}[width=0.76\linewidth]{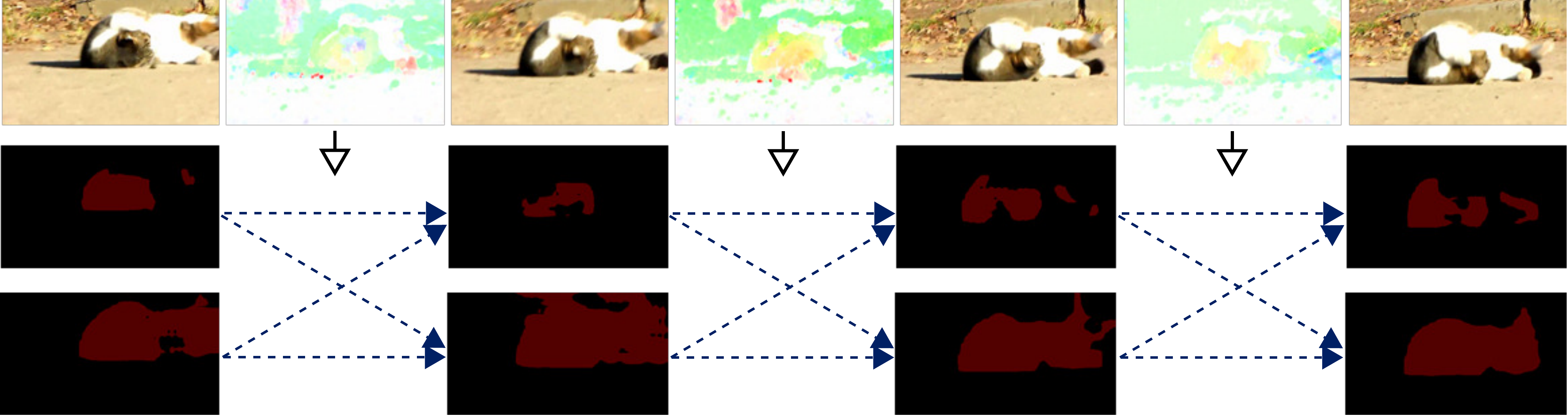} 
\put(-3,12.5){\footnotesize{$S^b$}}
\put(-3,3.5){\footnotesize{$S^o$}}
\put(15.3,19){\footnotesize{O.F.}}
\put(16.5,1.5){\footnotesize{$c(m_f,m_{f+1})$}}
\end{overpic}
\end{center}
\caption{Motion-consistent combination. Top row: consecutive frames of a video and color-coded optical flows (O.F.) between frames. Middle row: corresponding results of batch ($S^b$) model. Bottom row: corresponding results of online model ($S^o$). We compute the consistency $c(m_f,m_{f+1})$ between those results by following optical flow, then find the most consistent path of selected models. {\bf Best viewed in color.}}
\end{figure*}
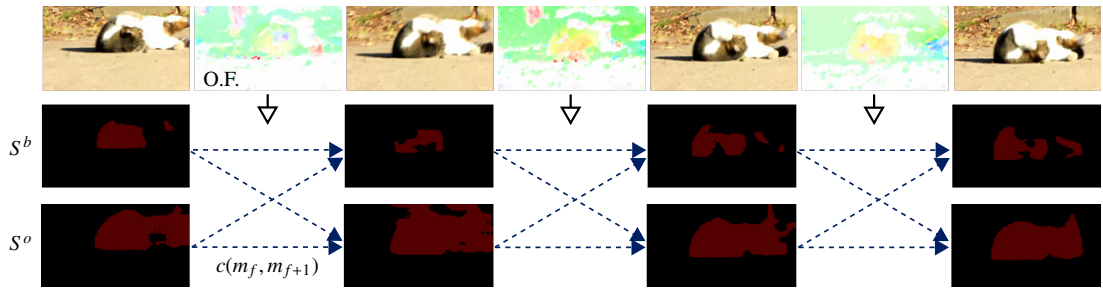
\subsection{Online}
The main difference between the batch and online frameworks is the generation of the self-adapting dataset. We adopt an online framework similar to object tracking using DCNN \citep{nam2015learning} because object tracking and online video object segmentation are closely related in that they should trace an object's appearance change while localizing object's region. \cite{nam2015learning} pre-trained domain-independent layers from training videos and collected two sets of frames (i.e., log-term and short-term sets) to fine-tune the model to be domain-specific. Similarly, to update the model periodically we collect and maintain two sets: ${\mathcal T}_l$ and ${\mathcal T}_s$ of frames instead of $\mathcal{G}$; ${\mathcal T}_l$ maintains $\tau_l$ globally-CE frames and ${\mathcal T}_s$ maintains $\tau_s$ locally-CE frames separately. ${\mathcal T}_l$ is implemented as a priority queue to collect globally-CE frames; ${\mathcal T}_s$ is a basic queue to deal with local variations. After collecting two sets of frames for a certain period $\tau_b$, we use both ${\mathcal T}_l$ and ${\mathcal T}_s$ as the self-adapting dataset to update the parameters of the model $\theta$. We iterate those procedures until the end of the video. The detailed procedures are described in Algorithm 2. Note that the first 11 lines in the algorithm are the same as the batch procedures.
\begin{algorithm}[t]
 \caption{AdaptNet-Online} \label{Alg1}
\begin{algorithmic}[1]
\STATE Given: DCNN model $\theta$, a set of weak labels $\mathcal{W}$
\STATE Local best confidence $d=0$
\STATE {\bf for} $f \in \mathcal{F}$ {\bf do}
\STATE ~~~~Initialize $G^g_f, G^l_f$ to "ignored" labels
\STATE ~~~~Compute $P({\bf x}|\theta)$ and $S=\arg\max_{\bf x}P({\bf x}|\theta)$
\STATE ~~~~Compute set $\mathcal{R}$ of connected components in $S$
\STATE ~~~~{\bf for} $R_k \in \mathcal{R}$ {\bf do}
\STATE ~~~~~~~~{\bf if} $S(i) \notin \mathcal{W}, i \in R_k$ {\bf then} {\bf continue}
\STATE ~~~~~~~~{\bf if} $C(R_k)>t_o$ {\bf then} Set $G^g_f(i)=S(i), \forall i \in R_k$
\STATE ~~~~~~~~Set $G^l_f(i)=S(i), \forall i \in R_k$
\STATE ~~~~Set $G^g_f(i)=G^l_f(i)=0, \forall i, s.t. P(x_i=bg|\theta)>t_b$
\STATE ~~~~{\bf if} $C(G^g_f)>0$ {\bf then} 
\STATE ~~~~~~~~{\bf if} $|{\mathcal T}_l|>\tau_l$ {\bf then} ${\mathcal T}_l$.dequeue
\STATE ~~~~~~~~~~~~ ${\mathcal T}_l$.enqueue($G^g_f$)
\STATE ~~~~{\bf if} $C(G^l_f)>d$ {\bf then}
\STATE ~~~~~~~~Update $t=f$ and $d=C(G^l_f)$
\STATE ~~~~{\bf if} $f$ mod $\tau_s = 0$ {\bf then}
\STATE ~~~~~~~~{\bf if} $G^g_t \notin {\mathcal T}_l$ {\bf then}
\STATE ~~~~~~~~~~~~{\bf if} $|{\mathcal T}_s|>\tau_s$ {\bf then} ${\mathcal T}_s$.dequeue
\STATE ~~~~~~~~~~~~${\mathcal T}_s$.enqueue($G^l_t$)
\STATE ~~~~~~~~Initialize $d=0$
\STATE ~~~~{\bf if} $f$ mod $\tau_b=0$ {\bf then}
\STATE ~~~~~~~~Finetune DCNN model $\theta$ to $\theta'$ using ${\mathcal T}_s \cup {\mathcal T}_l$
\STATE ~~~~~~~~Set $\theta \leftarrow \theta'$

\end{algorithmic}
\end{algorithm}

\subsection{Motion-Consistent Combination}

The batch algorithm generally works better than the online algorithm, because the former uses global update with larger pool of CE frames and has longer-range dependency than the latter; i.e., the batch algorithm makes its decision after processing the whole video. However, videos may exist in which the online algorithm shows better results for certain local frames. Thus we can combine the two results to improve the motion-consistency of segmentation by incorporating dense optical flow. We cast the combination as the problem of selecting the best model in every frame as follows: let $m_f\in\{S_f^{b},S_f^{o}\}, \forall f$ be a variable that selects a labeled result between batch $S_f^b$ and online $S_f^o$, and $c(m_f,m_{f+1})$ measure a consistency between two consecutive labeled frames. We can formulate the motion-consistent model selection problem as
\begin{equation*}
\argmax_{\bf m}\sum_f c(m_f,m_{f+1}),
\end{equation*}
where ${\bf m} = \{m_1, m_2,...,m_{|{\mathcal F}|}\}$ is the set of selected models. We measure the consistency $c(m_f,m_{f+1})$ by the overlap $o(m_f,m_{f+1})$ of object regions between consecutive labeled frames warped by following dense optical flow \citep{farneback2003two} (Fig. 3). Because the optical flow can be noisy we give a small preference $\epsilon$ for the transition from batch to batch result. That is,  
\begin{displaymath}
c(m_f,m_{f+1}) = \left\{ \begin{array}{ll}
o(m_f,m_{f+1})+\epsilon & \textrm{if } m_f=S_f^b \\ & \land ~ m_{f+1}=S_{f+1}^b \\
o(m_f,m_{f+1}) & \textrm{otherwise.}
\end{array} \right.
\end{displaymath}
Note that this problem can be easily solved using {\it dynamic programming}.

\begin{table*}[ht]
\centering
\caption{Intersection-over-union overlap on Youtube-Object-Dataset 2014 \citep{jain2014supervoxel}}
\begin{tabular}{lcccccccccc|c}
\Xhline{2\arrayrulewidth}
 & Aero & Bird & Boat & Car & Cat & Cow & Dog & Horse & Motor & Train & Avg. \\
\hline
Base-context & 0.808 & 0.642 & 0.627 & 0.746 & 0.622 & 0.646 & 0.670 & 0.414 & 0.570 & 0.607 & 0.635 \\
\hline
Base-front-end & 0.828 & 0.725 & 0.657 & 0.797 & 0.616 & 0.646 & 0.671 & 0.462 & 0.674 & 0.624 & 0.670 \\
\hline
SCF \citep{jain2014supervoxel} & \bf 0.863 & \bf 0.810 & 0.686 & 0.694 & 0.589 & 0.686 & 0.618 & 0.540 & 0.609 & 0.663 & 0.672 \\
\hline
Our-Unsupv-batch & 0.829 & 0.783 & 0.699 & 0.812 & 0.688 & 0.675 & 0.701 & 0.505 & 0.705 & 0.702 & 0.710 \\
Our-Weak-on & 0.819 & 0.774 & 0.686 & 0.791 & 0.676 & 0.680 & 0.710 & 0.540 & 0.693 & 0.679 & 0.705 \\
Our-Weak-batch & 0.830 & 0.788 & 0.708 & 0.817 & 0.688 & 0.685 & 0.732 & 0.589 & 0.711 & 0.718 & 0.727 \\
Our-Weak-comb & 0.830 & 0.788 & 0.710 & 0.817 & 0.713 & 0.696 & 0.732 & 0.595 & 0.711 & 0.718 & 0.731 \\
Our-Unsupv-CRF & 0.844 & 0.808 & 0.710 & 0.822 & 0.696 & 0.688 & 0.717 & 0.514 & 0.714 & 0.702 & 0.722 \\
Our-Weak-on-CRF & 0.837 & 0.794 & 0.690 & 0.797 & 0.694 & 0.690 & 0.726 & 0.553 & 0.704 & 0.673 & 0.716 \\
Our-Weak-batch-CRF & 0.844 & \bf 0.810 & 0.723 & \bf 0.827 & 0.698 & 0.700 & \bf 0.745 & 0.610 & \bf 0.722 & \bf 0.729 & 0.741 \\
Our-Weak-comb-CRF & 0.844 & \bf 0.810 & \bf 0.725 & \bf 0.827 & \bf 0.722 & \bf 0.709 & \bf 0.745 & \bf 0.611 & \bf 0.722 & \bf 0.729 & \bf 0.744 \\

\Xhline{2\arrayrulewidth}
\end{tabular}
\end{table*}
\subsection{Unsupervised Video}
We briefly mention our method for processing an unsupervised video. Our framework can be easily applied to unsupervised videos by bypassing line 8 in Algorithm 1. This deletion means that we do not care whether the class actually appears in the video, thus we set all the labels of CE regions even if the labels are incorrect. We found that most of the videos processed in this way show similar results to those of weakly-supervised video, because the labels of pixels determined with very high probability usually correspond to the correct labels. Nevertheless, a few exceptions that correspond to incorrect labels occur, which can degrade the model and decrease the accuracy compared with the weakly-supervised setting. 

\subsection{Post-processing}
Because the output of DCNN is insufficient to exactly delineate the object, we use the fully-connected CRF \citep{koltun2011efficient}. We simply use the output of DCNN for the unary term and use colors and positions of pixels for the computation of pairwise terms as \cite{chen2014semantic} did. Finally, we refine the label map through morphological operations (i.e., dilation and erosion). 

\section{Experiments}
\subsection{Implementation details}
We tested the `{\it front-end}' and `{\it context}' DCNN models pre-trained in \cite{yu2015multi} and observed that the {\it front-end} model shows better results than the {\it context} model under our setting on Youtube-Object-Dataset (Table 1). Thus we used the {\it front-end} model as our baseline model $\theta$. The {\it front-end} model is a modified version of the VGG-16 network \citep{simonyan2014very} and is extended for dense prediction. In practice, we resize each frame such that its long side is 500 pixels, then pad the frame by reflecting it about each image boundary to be 900$\times$900 pixels. We use threshold values $t_o=0.75$ and $t_b=0.8$; the value for background is set slightly higher than for foreground to leave room for additional foreground pixels (e.g., pixels around objects). We set the local period $\tau_b=30$ for both algorithms and $\tau_l=10$, $\tau_s=5$ for our online algorithm, and the small preference $\epsilon=0.02$ for the motion-consistent combination. For model update, we fine-tuned all the layers with dropout ratio of 0.5 for the last two layers and iterated for the number of frames in the self-adapting dataset with batch size of 1. We set the learning rate to 0.001, momentum to 0.9, and  weight decay to 0.0005. Due to the lack of a validation dataset, we use the fixed CRF parameters used in \citep{noh2015learning} during post-processing. Our implementation is based on the Caffe library \citep{jia2014caffe} equipped with an Nvidia GTX Titan X GPU.

\subsection{Evaluation}

\begin{table*}[t]
\centering
\caption{Intersection-over-union overlap on Youtube-Object-Dataset 2015 \citep{zhang2015semantic}}
\begin{tabular}{lcccccccccc|c}
\Xhline{2\arrayrulewidth}
 & Aero & Bird & Boat & Car & Cat & Cow & Dog & Horse & Motor & Train & Avg. \\
\hline
\citep{zhang2015semantic} & 0.758 & 0.608 & 0.437 & 0.711 & 0.465 & 0.546 & 0.555 & 0.549 & 0.424 & 0.358 & 0.541\\
\hline
Base-front-end& 0.786 & 0.727 & 0.632 & 0.866 & 0.583 & 0.657 & 0.632 & 0.403 & 0.635 & 0.626 & 0.655 \\
\hline
Our-Unsupv-batch & 0.791 & 0.766 & 0.681 & 0.876 & 0.679 & 0.711 & 0.656 & 0.432 & 0.689 & 0.650 & 0.693\\
Our-Weak-on & 0.795 & 0.773 & 0.665 & 0.883 & 0.641 & 0.691 & 0.684 & 0.488 & 0.658 & 0.652 & 0.693 \\
Our-Weak-batch & 0.794 & 0.786 & 0.685 & 0.870 & 0.667 & 0.738 & 0.734 & 0.567 & 0.694 & 0.672 & 0.721 \\
Our-Weak-comb & 0.794 & 0.786 & 0.684 & 0.870 & 0.668 & 0.738 & 0.737 & 0.580 & 0.694 & \bf 0.673 & 0.722 \\
Our-Unsupv-CRF & 0.808 & 0.791 & 0.695 & 0.879 & 0.683 & 0.729 & 0.669 & 0.441 & 0.690 & 0.651 & 0.704 \\
Our-Weak-on-CRF & \bf 0.813 & 0.796 & 0.672 & \bf 0.884 & 0.643 & 0.702 & 0.703 & 0.507 & 0.671 & 0.631 & 0.702 \\
Our-Weak--batch-CRF & 0.809 & \bf 0.809 & \bf 0.699 & 0.870 & 0.674 & \bf 0.756 & 0.751 & 0.580 & \bf 0.695 & 0.651 & 0.729\\
Our-Weak-comb-CRF & 0.809 & 0.807 & 0.698 & 0.870 & \bf 0.675 & \bf 0.756 & \bf 0.754 & \bf 0.595 & \bf 0.695 & 0.645 & \bf 0.730 \\

\Xhline{2\arrayrulewidth}
\end{tabular}
\end{table*}
We evaluate the proposed method on the Youtube-Object-Dataset \citep{jain2014supervoxel,zhang2015semantic} that contains subset of classes in the PASCAL VOC 2012 segmentation dataset, on which the baseline model is pre-trained. The Youtube-Object-Dataset was originally constructed by \cite{prest2012learning}. \cite{jain2014supervoxel} annotated pixel-level ground truth for every 10-th frame for the first shot of each video. The dataset consists of 126 videos with 10 object classes. Due to inconsistent numbers of annotations, \cite{zhang2015semantic} modified the dataset by resampling 100 frames (sampled every other frame) for each video and annotating missed frames for one in every 10 frames. We use those two versions of dataset to compare with the two existing methods respectively. We use the intersection-over-union overlap to measure the accuracy. 
 
The results on the dataset used in \citep{jain2014supervoxel} are shown in Table 1. We first report the accuracies of two baseline DCNN models proposed in \citep{yu2015multi} denoted by Base-front-end and Base-context in the table. Note that the Base-front-end model \citep{yu2015multi} showed higher accuracy (0.670) for the dataset under our setting than did the Base-context model (0.635), which attaches several context layers to the {\it front-end} model. It is interesting that the {\it front-end} model applied to each frame of unsupervised videos showed almost the same average accuracy as the results of SCF \citep{jain2014supervoxel}, which is built based on semi-supervised video. 

\begin{figure*}[t]\centering
\begin{tabular}{ccccc}
Aeroplane & Bird & Boat & Car & Cat \\
\includegraphics[width=1.26in]{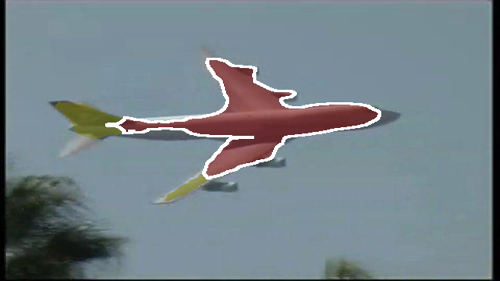}&
\includegraphics[width=1.26in]{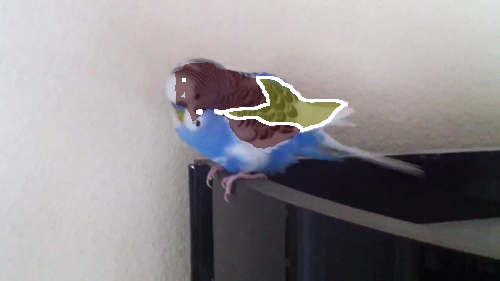}&
\includegraphics[width=1.26in]{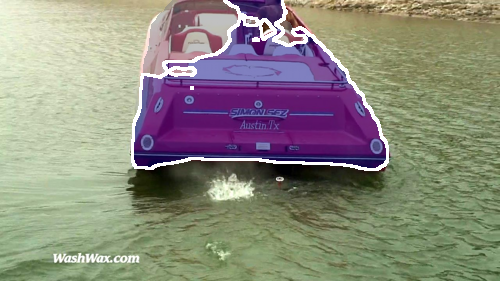}&
\includegraphics[width=1.26in]{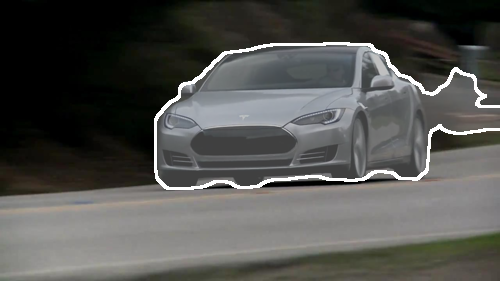}&
\includegraphics[width=1.26in]{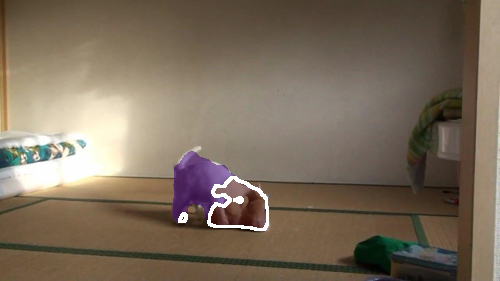}\\
\includegraphics[width=1.26in]{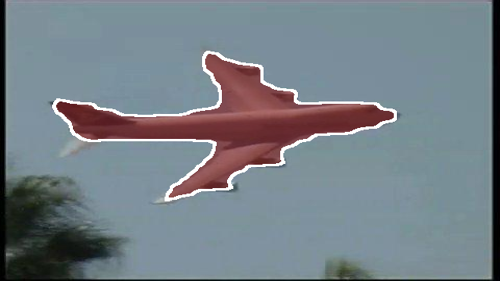}&
\includegraphics[width=1.26in]{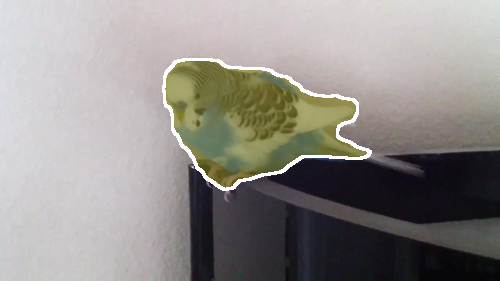}&
\includegraphics[width=1.26in]{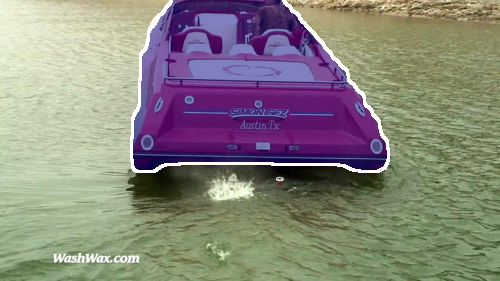}&
\includegraphics[width=1.26in]{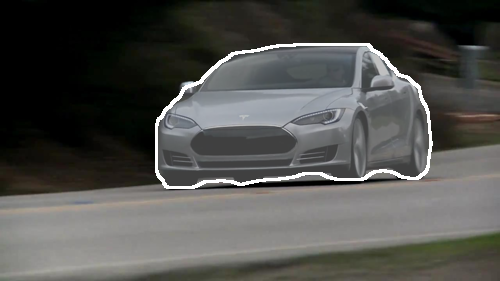}&
\includegraphics[width=1.26in]{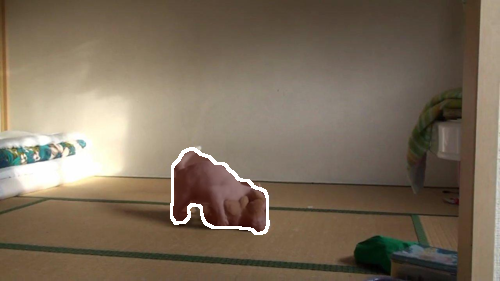}\\
\includegraphics[width=1.26in]{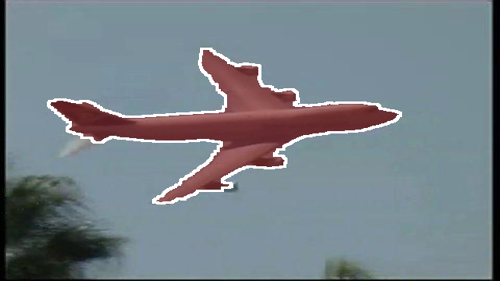}&
\includegraphics[width=1.26in]{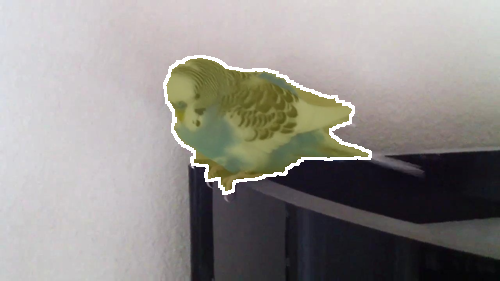}&
\includegraphics[width=1.26in]{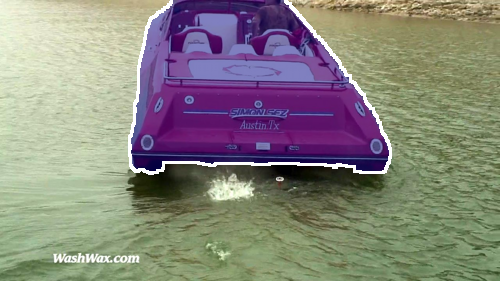}&
\includegraphics[width=1.26in]{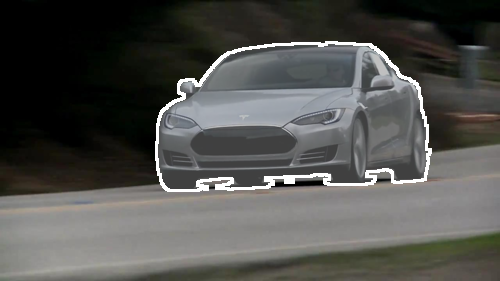}&
\includegraphics[width=1.26in]{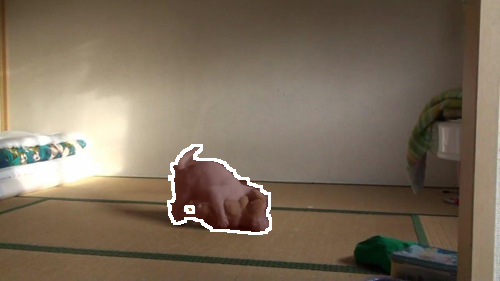}\\
Cow & Dog & Horse & Motorbike & Train\\
\includegraphics[width=1.26in]{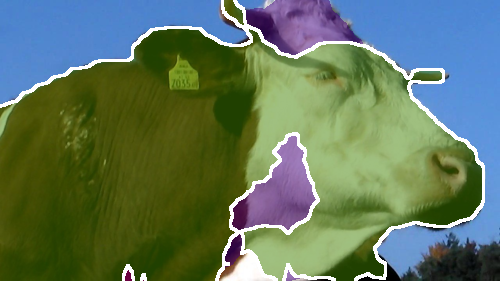}&
\includegraphics[width=1.26in]{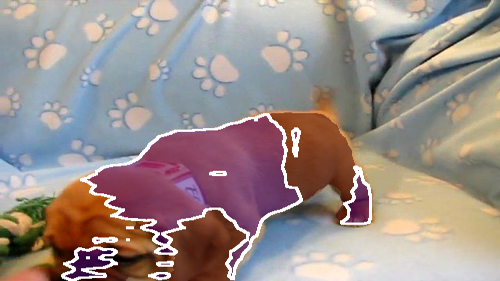}&
\includegraphics[width=1.26in]{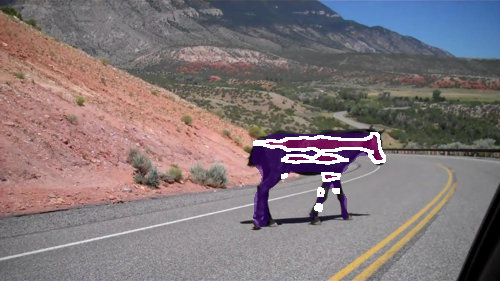}&
\includegraphics[width=1.26in]{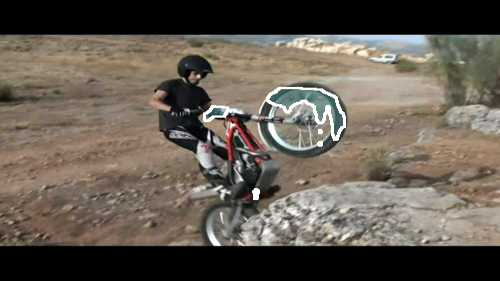}&
\includegraphics[width=1.26in]{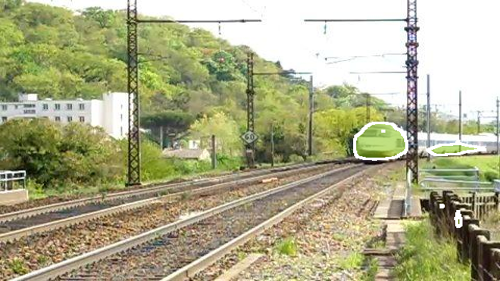}\\
\includegraphics[width=1.26in]{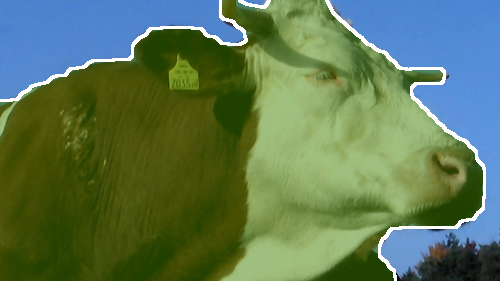}&
\includegraphics[width=1.26in]{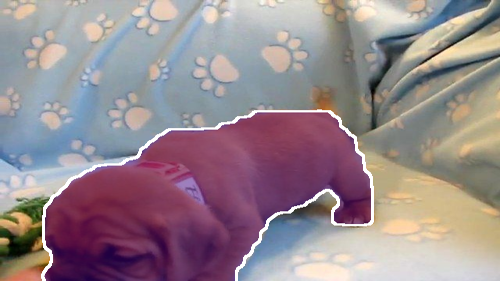}&
\includegraphics[width=1.26in]{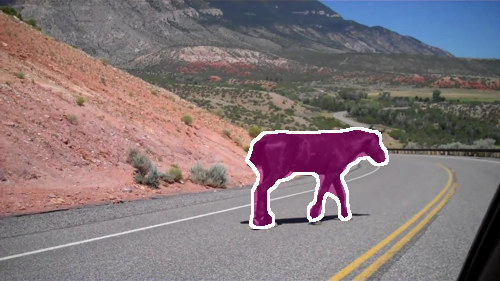}&
\includegraphics[width=1.26in]{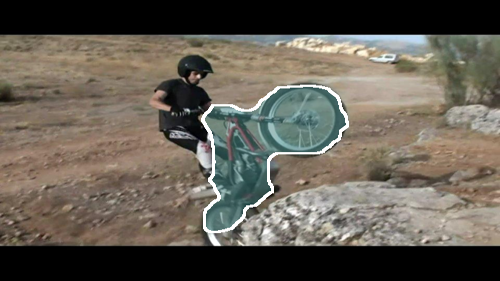}&
\includegraphics[width=1.26in]{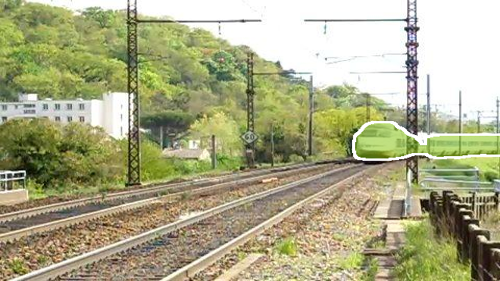}\\
\includegraphics[width=1.26in]{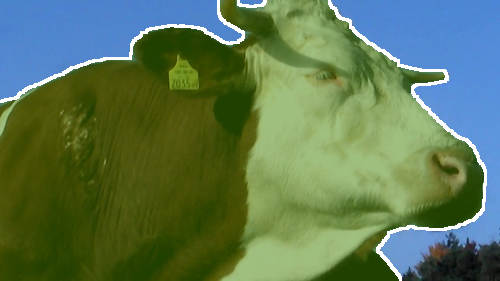}&
\includegraphics[width=1.26in]{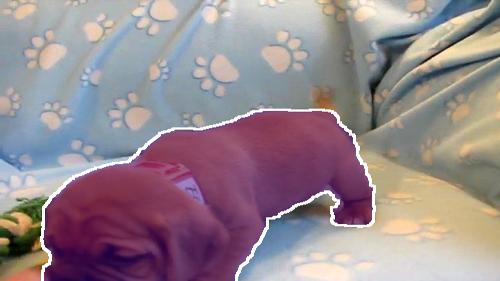}&
\includegraphics[width=1.26in]{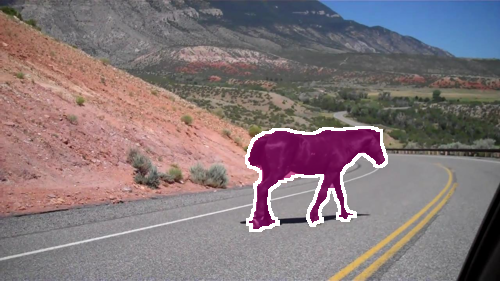}&
\includegraphics[width=1.26in]{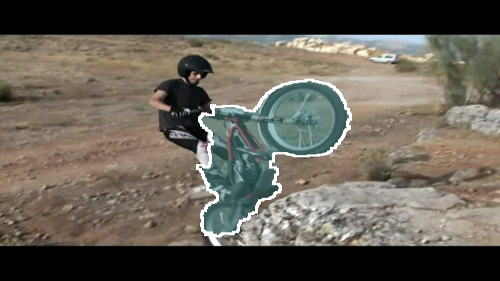}&
\includegraphics[width=1.26in]{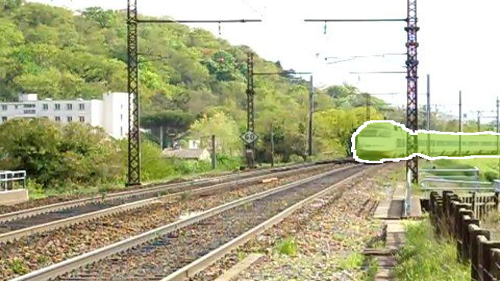}\\

\end{tabular}
\caption{Representative results of proposed method compared with baseline model. The results of Top: Base-front-end, Middle: Our-Weak-comb, Bottom: Our-Weak-comb-CRF. Semantic labels are overlaid on images with different colors corresponding to different class labels. We only highlight the boundary of correct class. {\bf Best viewed in color.}} 
\label{fig:qualitative}
\end{figure*}

Our method with weak-supervision (Our-Weak-comb-CRF) further improved the accuracy to 0.744, which exceeds that of SCF (average 0.672). Our online algorithm under weak-supervision (Our-Weak-on) improved the accuracy by 3.5\% and the batch algorithm (Our-Weak-batch) improved it by 5.7\% which is better than the online algorithm due to the global update. Note that our motion-consistent combination (Our-Weak-comb) achieved small improvement on a few classes by selecting results that are more motion-consistent, which shows temporally consistent video segmentation result. The great improvement of our model over the baseline model mostly originates from the correction of confusing parts of frames by the newly-updated model based on CE frames. Some representative results of ours are shown in Figure \ref{fig:qualitative}\footnote{More video results and the datasets we used are available at the project webpage: {\it https://seongjinpark.github.io/AdaptNet/}}. We also report our results without any supervision (Our-Unsupv-batch) explained in Sec 3.4 to show the efficiency of our method. It achieved slightly lower accuracy (0.710) than the algorithm with weak-supervision, but had 3.8\% higher average accuracy than SCF. We report only the batch algorithm for unsupervised videos because the online algorithm is more vulnerable than the batch algorithm to incorrect labels, especially for the short videos that the dataset includes. Post-processing increased the accuracy by about $1.1\sim1.4$\%. Our result achieved state-of-the-art overlap accuracy for all classes except `aeroplane'. We observed that the slightly inferior accuracy on this class occurs mainly because our method gives less accurate object boundary for the class. This problem occurs because we used the same CRF parameters for all classes as we could not cross-validate the parameters for each class, whereas SCF is given manual delineation of the object in the first frame.

In Table 2, we also report our results on the dataset used in \cite{zhang2015semantic} to compare with the framework that uses models pre-trained on images to segment weakly-supervised video. \cite{zhang2015semantic} constructed the dataset (Youtube-Object-Dataset 2015) by modifying the Youtube-Object-Dataset 2014. Because neither the dataset nor source code is provided by the authors, we manually built the dataset by following the procedures explained in the paper. Our results in Table 2 show tendencies similar to those in Table 1, from the online to the combined model that greatly improves the baseline model. In this analysis the difference between the accuracies of our model and that of existing method was much larger than in Table 1 because \cite{zhang2015semantic} used a conventional object detector on weakly-supervised video, whereas we use a semantic segmentation model based on DCNN. 

\subsection{Limitation}
The limitation of our method occurs when a video does not meet our assumption that at least one frame has a correct label or that at least one object region that corresponds to the pre-trained object classes is estimated. Such cases mostly occur due to very small size of objects in an image. The absence of a frame to improve the other frames yields the same result as the baseline model. We plan to consider these problems in our future work.

\section{Conclusion}
We proposed a novel framework for video semantic object segmentation that adapts the pre-trained DCNN model to the input video. To fine-tune the extensively-trained model to be video-specific, we constructed a self-adapting dataset that consists of several frames that help to improve the results of the UE frames. In experiments the proposed method improved the results by using the fine-tuned model to re-estimate the misclassified parts. It also achieved state-of-the-art accuracy by a large margin. We plan to extend the framework for semi-supervised video to increase the accuracy. We also expect that the efficient self-adapting framework can be applicable to generate a huge accurately-labeled video dataset, and thus be used to progress image semantic segmentation.

\bibliographystyle{model2-names}
\bibliography{refs}

\end{document}